\documentclass{jpconf}
\usepackage{amsmath}
\usepackage{multirow,rotating}
\usepackage{iopams}
\usepackage{amsfonts,amssymb}
\usepackage{amsthm}
\usepackage[table,dvipsnames]{xcolor}
\usepackage{times}
\usepackage[T1]{fontenc}
\usepackage[utf8]{inputenc}
\usepackage{mathptmx}
\usepackage{graphicx}
\usepackage{cite}
\usepackage{hyperref}

\newcommand{\raiseline}[1]{\raisebox{1.5ex}[-1.5ex]{#1}}
\newcommand{\pfend}{\hfill\ensuremath{\Box}}
\newcommand{\tiff}{if and only if \ }
\newcommand{\z}{\emptyset}
\newcommand{\klam}[1]{\ensuremath{\langle #1 \rangle}}
\newcommand{\set}[1]{\ensuremath{\{#1\}}}
\newcommand{\card}[1]{\ensuremath{\lvert #1 \rvert}}
\newcommand{\X}{\ensuremath{\mathcal{X}}}
\newcommand{\Y}{\ensuremath{\mathcal{Y}}}
\newcommand{\upp}{\ensuremath{\operatorname{\mathit{Upp}}}}
\newcommand{\low}{\ensuremath{\operatorname{\mathit{Low}}}}
\newcommand{\df}{\ensuremath{:=}}

\theoremstyle{plain}
\newtheorem{theorem}{Theorem}[section]
\newtheorem{lemma}{Lemma}[section]

\theoremstyle{definition}
\newtheorem{example}{Example}

\numberwithin{equation}{section}

\begin{document}

\title{Confusion matrices and rough set data analysis}

\author{Ivo D\"{u}ntsch%
\footnote{The ordering of authors is alphabetical and equal authorship is implied.}
\footnote{Permanent address: Dept. of Computer Science, Brock University, St Catharines, Canada}
$^{3}$, G{\"u}nther Gediga$^{1~4}$
}
\address{$^3$ College of Mathematics and Informatics,  Fujian Normal University,  Fuzhou, China}
\address{$^4$ Institut f{\"u}r Evaluation  und Marktanalysen, Brinkstr.~19, 49143 Jeggen, Germany}
\ead{D.Ivo@fjnu.edu.cn,gediga@eval-institut.de}

%\maketitle

\begin{abstract}
\noindent A widespread approach in machine learning to evaluate the quality of a classifier is to cross -- classify predicted and actual decision classes in a \emph{confusion matrix}, also called \emph{error matrix}. A classification tool which does not assume distributional parameters but only information contained in the data is based on rough set data model which assumes that knowledge is given only up to a certain granularity. Using this assumption and the technique of confusion matrices, we define various indices and  classifiers based on \emph{rough confusion matrices}.
\end{abstract}

\section{Introduction}

In pattern recognition and other disciplines of machine learning, the sum of the diagonal elements of a confusion matrix is widely used to measure the success of a classification based on an algorithm or human observation in comparison with a gold standard (or ``true'' measurement) such as classification by an expert. The main idea is that an algorithm (or an observer) forms its own hidden equivalence classes of the data, and is forced to assign the classes to the categories given by the gold standard. The underlying model may be one of a plethora of existing techniques see e.g. \cite{nvipt17,hand_supervised_2005,Caelen2017}. The question may be asked, whether such an index is valid for determining the quality of a classifier: Since we approximate sets, namely, decision classes, one should use a theory of set approximation such as the rough set approach to investigate this question.

In a first step we find a connection of a rough set decision system and a resulting confusion matrix. We derive several approximations of upper and lower bounds of the classes given by the gold standard; additionally, we consider the standard indices of rough set analysis for the coverage. Owing to lack of space we shall only indicate the procedures, and  detailed results and proofs will appear elsewhere.

\section{Definitions and notation}

Throughout, $U$ denotes a finite nonempty set with $n$ elements. Given a set $\Y = \set{Y_1, \ldots, Y_k}$ of decision classes, a \emph{classifier} is a mapping $f:U \to \Y$ which predicts the class membership of an element of $U$ in a decision class. The predicted and true values of class membership can be cross--classified and counted in a \emph{confusion matrix}. If success of a classifier is measured by error rate, confusion matrices may be used to analyse and to compare classifiers. A widely used confusion matrix of dimension two is shown in Table \ref{tab:cont}, and a general confusion matrix is shown in Table \ref{tab:rule}. An entry $\klam{\hat{Y}_i, Y_j} = n_{ij}$ in the matrix is the number of elements of $Y_j$ which are predicted to be in $Y_i$; in particular, $\sum\set{n_{ii}: 1 \leq i \leq k}$ is the number of correctly classified elements.

\begin{table}[!htb]
\begin{minipage}[t]{0.33\textwidth}
\caption{A 2--class confusion matrix}\label{tab:cont}
\centering
{\scriptsize
\begin{tabular}{cccc}
  & & \multicolumn{2}{c}{\bfseries True value} \\
& & $P$ & $N$ \\ \hline \\
 &  & True & False \\
  \multirow{3}{*}{\rotatebox{90}{\parbox{1.1cm}{\bfseries\centering Predicted value}}}
 &\raiseline{$\hat{P}$}& Positive & Positive \\
 & & False & True \\
 &\raiseline{$\hat{N}$}& Negative & Negative
 \end{tabular}
 }
\end{minipage}
\begin{minipage}[t]{0.63\textwidth}
\caption{A general confusion matrix}\label{tab:rule}
\centering
{\scriptsize
\begin{tabular}{cccccccc}
&& & \multicolumn{4}{c}{\bfseries True value} \\
& &  $Y_1$ & ... & $Y_i$ & ... & $Y_k$ & Sum \\ \hline \\
\multirow{5}{*}{\rotatebox{90}{\parbox{1.1cm}{\bfseries\centering Predicted value}}}
&$\hat Y_1$&  $n_{11}$  & ... &  $n_{1i}$   & ... & $n_{1k}$    & $n_{1\bullet}$   \\
&... &  ...        & ... &  ...        & ... & ... & ... \\
&$\hat Y_i$&  $n_{i1}$  & ... &  $n_{ii}$   & ... & $n_{ik}$    & $n_{i\bullet}$   \\
&... &  ...        & ... &  ...        & ... & ... & ... \\
&$\hat Y_k$&  $n_{k1}$  & ... &  $n_{ki}$   & ... & $n_{kk}$    & $n_{k\bullet}$   \\ \hline
&Sum &  $n_{\bullet 1}$  & ... &  $n_{\bullet i}$   & ... & $n_{\bullet k}$ & n \\ \hline
\end{tabular}
}
\end{minipage}
\end{table}

The philosophy of rough sets is based on the assumption that knowledge of the world depends on the granularity of representation \cite{paw82}. Mathematically, granularity may be expressed by an equivalence relation $\theta$ on a nonempty finite set $U$, up to the classes of which membership in a subset of $U$ can be determined. For rough approximation, two operators are defined on $2^U$ in the following way: Let $\X \df \set{X_1, \ldots, X_m}$ be the set of equivalence classes of $\theta$. If $Y \subseteq  U$,  then,
{\small
\begin{xalignat}{2}
\low_\X(Y) &\df \bigcup \set{Z \in \X: Z \subseteq Y}, &&\text{\emph{Lower approximation},}\label{def:low}\\
\upp_\X(Y) &\df \bigcup \set{Z \in \X: Z \cap Y \neq \z},  &&\text{\emph{Upper approximation}.} \label{def:upp}.
%\bnd_\X(Y) &\df \upp_\X(Y) \setminus \low_\X(Y)  &&\text{\emph{Boundary}}.
\end{xalignat}
}

The main data type of the rough set approach are \emph{decision systems} which are closely related to relational data tables with an added decision attribute. An example is shown in Table \ref{decsys}; there, the object set $U$ contains six elements, there are four independent attributes, and one decision attribute $d$.

\begin{table}[!ht]
\begin{minipage}[t]{0.49\textwidth}
\vspace{0mm}
\caption{A decision system}\label{decsys}
\vspace{2mm}
\centering
{\scriptsize
\begin{tabular}{|c|c|c|c|c||c|} \hline
Type & Price & Guarantee & Sound & Screen & d \\ \hline\hline
1 &  high & 24 months & Stereo & 76 & high \\ \hline
2& low & 6 months & Mono & 66 & low \\ \hline
3& low & 12 months & Stereo & 36 & low \\ \hline
4& medium & 12 months & Stereo & 51 & high \\ \hline
5& medium & 18 months & Stereo & 51 & high \\ \hline
6& high& 12 months & Stereo & 51 & low \\ \hline
\end{tabular}
}
\end{minipage}
\begin{minipage}[t]{0.49\textwidth}
\vspace{0mm}
\caption{A granule frequency matrix}\label{tab:gran}
\centering
{\scriptsize
\begin{tabular}{ccccccc}
& \multicolumn{6}{c}{Decision classes \Y} \\
Granules \X   &  $Y_1$ & ... & $Y_i$ & ... & $Y_k$ & Granule size \\ \hline
$X_1$&  $c_{11}$  & ... &  $c_{1i}$   & ... & $c_{1k}$    & $c_{1}$   \\
... &  ...        & ... &  ...        & ... & ... & ... \\
$X_j$&  $c_{j1}$  & ... &  $c_{ji}$   & ... & $c_{jk}$    & $c_{j}$   \\
... &  ...        & ... &  ...        & ... & ... & ... \\
$X_m$&  $c_{m1}$  & ... &  $c_{mi}$   & ... & $c_{mk}$    & $c_{m}$   \\ \hline
Decision class size  &  $n_{1}$  & ... &  $n_{i}$   & ... & $n_{k}$ & n \\ \hline
\end{tabular}
}
\end{minipage}
\end{table}

For simplicity of notation, we suppose that an attribute $a$ is a mapping from $U$ to the set $V_a$ of values of $a$. Each set $Q$ of independent  attributes gives rise to an equivalence relation $\theta_Q$ on $U$ by setting $x \theta_Q y$ \tiff $a(x) = a(y)$ for all $a \in Q$. Similarly, the decision attribute $d$ induces an equivalence relation $\theta_d$, the classes $\Y \df \set{Y_1, \ldots, Y_k}$ of which are called \emph{decision classes}. We cross--classify the classes of $\theta$ with the decision classes in a \emph{granule frequency matrix}, see Table \ref{tab:gran}; there, $c_j = \card{X_j}$, $n_i = \card{Y_i}$, and $c_{ij} = \card{X_i \cap Y_j}$. Furthermore, we introduce the following parameters for each decision class $Y_i$:
\begin{gather}\label{pars}
n_i \df \card{Y_i}, \ nl_i \df \card{\low(Y_i)}, \ nu_i \df \card{\upp(Y_i)}.
\end{gather}
Consider the vector $\vec{X}_i = \klam{c_{ij}: 1 \leq j \leq k}$ belonging to granule $X_i$.
%
%Since $\Y$ is a partition of $U$ and $X_i \neq \z$, it follows that $c_{ij} \neq 0$ for at least one $1 \leq j \leq k$.
%
If $\vec{X}_i$  contains only one non--zero entry, we call the granule \emph{deterministic}. In this case, $X_i \subseteq Y_j$ and prediction based on $X_i$ is perfect. Otherwise, the granule is called \emph{indeterministic}. A subset $Y$ of $U$ is called \emph{definable}, if it is a union of elements of \X.

A major aim of rough set data analysis is to decide (or estimate) membership of an element $x$ of $U$ in a decision class using the knowledge given by a set $Q$ of attributes, in particular, how well the decision classes can be approximated by the knowledge obtained from a partition induced by $Q$. Note that we can define a partial classifier $f_r$ as follows: If $D = \bigcup\set{X_i: X_i \text{ is a deterministic class}}$, then each $x \in D$ is correctly classified (and these are the only ones). Thus we can set $f_r(x) = x$ for all $x \in D$. If $x \in X_i$ and $x \not\in D$, then the rough method assigns $x$ to one ore more upper approximations of decision classes. In this sense, rough approximation is not a point estimate. With some abuse of language, we call $f_r$ a rough classifier.

In the sequel, we suppose that $\X = \set{X_1, \ldots, X_m}$ is the set of classes of a fixed equivalence relation $\theta$ on $U$, called \emph{granules}, and $\Y = \set{Y_1, \ldots, Y_k}$ is a set of decision classes; to avoid trivialities we assume that $k > 1$. Lower and upper approximations are taken with respect to $\X$, and we shall omit the indices in the approximation functions. We shall write $Z = Z_1 \uplus \ldots \uplus Z_r$ if $Z = Z_1 \cup \ldots \cup Z_r$, and the sets $Z_i$ are pairwise disjoint. At times, we are only interested whether the entry in a cell is $0$ or not. To this end, we introduce an indicator function $Ind: \mathbb{N} \to \set{0,1}$ defined by
\begin{gather}\label{def:ind}
Ind(b) \df
\begin{cases}
0, &\text{if } b
 = 0, \\
1, &\text{otherwise},
\end{cases}
\end{gather}

For the basic philosophy and tools of the rough set method the reader is invited to consult \cite{dg_noninv2}. For recent developments and more advanced methods the overview \cite{ns13} is an excellent source.

\section{Rough confusion matrices}

According to the rough set philosophy, we can only distinguish elements of $U$ up to equivalence with respect to $\theta$, hence, we must have $f(x) = f(y)$ for any classifier $f$ whenever $x$ and $y$ are in the same granule. Thus, with some abuse of language, we call a function $f: \X \to \Y$ a \emph{(rough) classifier}.
The meaning of the classifier $f$ is that each element of $X_i$ is predicted to be in $f(X_i)$. Thus, we obtain the predictor sets
\begin{gather}\label{def:hat}
\hat{Y}_i \df \bigcup f^{-1}(Y_i) = \bigcup\set{X_s: f(X_s) = Y_i}.
\end{gather}
If $\hat{Y}_i = \z$, then no element of $U$ is predicted to be in $Y_i$ by any class $X_s$ using $f$.
%
%Since each nonempty $\hat{Y}_j$ is a union of equivalence classes of $\X$, and $f$ is a function, the collection $\set{f^{-1}(Y_j): 1 \leq j \leq k, f^{-1}(Y_j) \neq \z}$ is a partition of $\X$, and therefore, keeping in mind that $\X$ partitions $U$, it follows that $\bigcup \set{f^{-1}(Y_j): 1 \leq j \leq k, f^{-1}(Y_j) \neq \z} = U$.
%
The \emph{(rough) confusion matrix} of the classifier $f$ has dimension $k \times k$, row labels $\hat{Y}_i$, column labels $Y_j$ and, for $1 \leq i,j \leq k$, the entries
\begin{align}\label{def:cell}
n_{ij} &\df
\begin{cases}
\sum\set{c_{sj}: f(X_s) = Y_i}, &\text{if }f^{-1}(Y_i) \neq \z, \\
0, &\text{otherwise.}
\end{cases}
\end{align}
Thus, $n_{ij} = \sum_{f(X_s) = Y_i} \card{X_s \cap Y_j}$. Since \X\ is a partition of $U$, $n_{ii} \leq \card{Y_i}$ for all $1 \leq i \leq k$.
%Thus, $c_{sj} = \card{X_s \cap Y_j}$ implies $n_{ij} = \sum_{f(X_s) = Y_i} \card{X_s \cap Y_j}$.

The rough confusion matrix can be obtained in several steps:

\begin{enumerate}
\item Write the granule frequency matrix $\mathfrak M$ obtained from $\X$ and $\Y$ as in Table \ref{tab:gran}.
\item Relabel the rows of $\mathfrak M$ by $f(X_i)$ by replacing $X_i$ with $f(X_i)$.
\item Aggregate the frequencies of the rows with the same label according to \eqref{def:cell}. If $f^{-1}(Y_j) = \z$, fill the row labeled $\hat{Y}_j$ with $0$s.
\item Sort the rows according to indices of their labels. The result has the form shown in Table \ref{tab:rule}.
\end{enumerate}

\begin{example}\label{ex:conf1}

We shall use the decision system of Table \ref{decsys}. Let $\theta$ be the equivalence relation generated by the attributes \emph{Price} and \emph{Screen}. The partition generated by $\theta$ has the classes
\begin{gather*}
X_1 = \set{1,6}, \ X_2 = \set{2}, \ X_3 = \set{3}, \ X_4 = \set{4,5},
\end{gather*}
and the decision classes
\begin{gather*}
Y_1 = \set{1,4,5}, \ Y_2 = \set{2,3,6}.
\end{gather*}
We define $f: \X \to \Y$ by $f(X_1) = f(X_4) = Y_1$, and $f(X_2) = f(X_3) = Y_2$. The construction process is shown in Tables \ref{tab:gfm1}, \ref{tab:rl1}, and \ref{tab:cm1}.
\begin{table}[!htb]
\begin{minipage}[t]{0.33\textwidth}
\vspace{0mm}
\caption{The granule freq. matrix}\label{tab:gfm1}
{\scriptsize
$$
\begin{array}{cccc}
& Y_1 & Y_2  &\text{Sum}\\
X_1 & 1 & 1 &2 \\
X_2 & 0 & 1 &1 \\
X_3 & 0 & 1 &1 \\
X_4 & 2 & 0 & 2 \\
\text{Sum} & 3 & 3 & 6
\end{array}
$$
}
\end{minipage}
\begin{minipage}[t]{0.32\textwidth}
\vspace{0mm}
\caption{The relabeled matrix}\label{tab:rl1}
{\scriptsize
$$
\begin{array}{cccc}
& Y_1 & Y_2  &\text{Sum}\\
Y_1 & 1 & 1 &2 \\
Y_2 & 0 & 1 &1 \\
Y_2 & 0 & 1 &1 \\
Y_1 & 2 & 0 & 2 \\
\text{Sum} & 3 & 3 & 6
\end{array}
$$
}
\end{minipage}
\begin{minipage}[t]{0.32\textwidth}
\vspace{0mm}
\caption{The confusion matrix}\label{tab:cm1}
{\scriptsize
$$
\begin{array}{ccccc}
& Y_1 & Y_2 & \text{Sum}\\
\hat{Y}_1 & 3 & 1 &4 \\
\hat{Y}_2 & 0 & 2 &2 \\
\text{Sum} & 3 & 3& 6
\end{array}
$$
}
\end{minipage}
\end{table}
Note that $f$ classifies five of the six elements of $U$ correctly, so that its success ratio is $\frac{5}{6}$, where as $\gamma = \frac{4}{6}$.
\pfend
\end{example}

According to the rough set philosophy, the set $\low(Y_i)$ approximates the diagonal set  $\hat{Y}_i\cap Y_i$. The optimal approximation would be $\low(Y_i) = \hat{Y}_i\cap Y_i$ with $|\hat{Y}_i\cap Y_i|=n_{ii}$; in this case, $Y_i$ is deterministic with respect to \X. Without knowledge of the source information system, but given the resulting confusion matrix, we obtain only $|\low(Y_i)| \leq n_{ii}$. Similarly, it is easy to see that $|\upp(Y_i)| \geq n_{i.}+n_{.i}-n_{ii}$.

Two statistics are of importance in the rough set literature: The \emph{rough approximation quality}
 is the weighted sum
\begin{gather}\label{gamma}
\gamma = \sum_{i=1}^k \frac{n_i}{n} \cdot p_i,
\end{gather}
and the \emph{accuracy of approximation} of the decision class $Y_i$ is defined by the index
\begin{gather}\label{alpha}
\alpha_i = \frac{nl_i}{nu_i} = p_i \cdot p^i.
\end{gather}
Here, $p_i \df \frac{nl_i}{n_i}$ and $p^i \df \frac{n_i}{nu_i}$ are precision indices \cite{dg_gamma}. The measure $\alpha_i$ is the maximal (best possible) value for the approximation quality of the set $Y_i$ of an information system which produces the observed confusion matrix.

Note that $\gamma$ and the upper bound weighted mean value
\begin{gather}
\alpha \df \frac{\sum_i (n_{i \bullet} + n_{\bullet i}) \cdot \alpha_i}{\sum_i n_{i \bullet}+n_{\bullet i}-n_{ii}}
\end{gather}
of the $\alpha_i$ are linked by a strictly monotone transformation, since
\begin{gather}\label{alphamean}
\alpha = \frac{\sum_i (n_{i \bullet} + n_{\bullet i}) \cdot \alpha_i}{\sum_i n_{i \bullet}+n_{\bullet i}-n_{ii}} =  \frac{\sum_i n_{ii}}{\sum_i n_{i \bullet}+n_{\bullet i}-n_{ii}} = \frac{\gamma}{2-\gamma}.
\end{gather}
Therefore, they are interchangeable as a measure of overall approximation quality.

The $\alpha$ -- accuracy is connected to the confusion matrix (and not to the underlying information system) by $\alpha_i = \frac{nl_i}{nu_i} = \frac{n_{ii}}{n_{i \bullet}+n_{\bullet i}-n_{ii}} $.
As $\alpha$ is a weighted mean of the $\alpha_i$ and $\gamma$ is a strictly monotone function of $\alpha$, we observe that upper confusion $\gamma$ and upper confusion $\alpha$ are maximal as well.

\section{Refining the rough classifier}

Thus far, we have put no restrictions on the classifier function $f$. In order to bring the concept closer to rough sets, and use more of the available information, we shall suppose in the sequel that a rough classifier satisfies the condition
\begin{gather}\label{def:rule}
X_i \cap f(X_i) \neq \z.
\end{gather}
\eqref{def:rule} implies that at least one element of $X_i$ is classified correctly by $f$. Furthermore,
\begin{lemma}\label{lem:cond}
\begin{enumerate}
\item If $X_i \subseteq Y_j$, then $f(X_i) = Y_j$.
\item $\low(Y_j) \subseteq  \hat{Y}_j$.
\item If $n_{ii} = 0$, then $n_{ij} = 0$ for all $1 \leq j \leq k$.
\end{enumerate}
\end{lemma}

Our first task is to approximate $nl_j = \card{\low(Y_j)}$. To this end, we first consider $n_j^* \df n_{jj}$. The cell $n_{jj}$ counts, in particular, the cardinality of the deterministic granules contained in $Y_j$, and thus, $nl_j \leq n_j^*$. We can further remove certain entries, and define $nl_j^{**} \df  n_{jj} - Ind\left(\sum_{j\neq i} n_{ji}\right)$. Using Lemma \ref{lem:cond} it is not hard, if somewhat tedious, to show the relationships among these indices:

\begin{theorem}\label{thm:nl}
Let $1 \leq j \leq k$. Then,
\begin{gather}
nl_j \leq nl_j^{**} \leq nl_j^{*} \leq \card{Y_j}.
\end{gather}
\end{theorem}
Not all of these inequalities need to hold if $f$ does not satisfy \eqref{def:rule}.

Turning to upper approximations, we first observe that \eqref{def:rule} is equivalent to $X_i \subseteq \upp(f(X_i))$ by \eqref{def:upp}, and thus, $\hat{Y}_j$ is a lower bound of the rough upper approximation of $Y_j$, i.e. $\card{\hat{Y}_j} \leq nu_j$. This can be sharpened as follows: Set
\begin{gather*}
nu_j^* \df n_{jj} + \sum_{i\neq j} (n_{ij} + n_{ji})= \sum_{i\neq j} n_{ij} + \sum_{i} n_{ji}.
\end{gather*}
A moment's reflection shows that $\sum_{i} n_{ji}$ adds all the cells in the partial granule frequency matrix spanned by the rows $X_i$ where $f(X_i) = Y_j$, and $\sum_{i\neq j} n_{ij}$ adds the entries $c_{ij}$, where $X_i \cap Y_j \neq \z$ and $f(X_i) \neq Y_j$.

If $n_{ij} \neq 0$, then $n_{ii} \neq 0$ by Lemma \ref{lem:cond}, and therefore, there is some $X_s$, such that $f(X_s) = Y_i$ and $X_s \cap Y_j \neq \z$, i.e. $X_s \subseteq \upp(Y_j)$. Therefore,  if $n_{ij} \neq 0$, there is at least one additional element which is in $\upp(Y_j)$. Hence, we obtain a sharper bound by setting $nu_j^{**} \df nu^*_{j} + \sum_{i\neq j} Ind(n_{ij}))$. Altogether, this leads to the following result:
\begin{theorem}\label{thm:nu}
Let $1 \leq j \leq k$. Then,
\begin{gather}
\card{Y_j} \leq nu^*_j \leq nu^{**}_j \leq nu_j.
\end{gather}
\end{theorem}

Arguably, the simplest classifier that satisfies \eqref{def:rule} is a \emph{maximal row classifier} $f_{mrc}$ defined as follows: Consider a granule frequency matrix shown in Table \ref{tab:gran}. For each $1 \leq i \leq m$ choose some $1 \leq j \leq k$ such that $c_{ij}$ is maximal in $\set{c_{i1}, \ldots, c_{ik}}$. Such $j$ always exists, but the choice need not be unique. Then, set $f_{mrc}(X_i) \df Y_j$. The classifier $f_{mrc}$ satisfies \eqref{def:rule}, and it is well compatible with the rough set philosophy in using only information supplied by the data.
%The classifier of Example \ref{ex:conf1} has this form.
%

By definition, $X_i \subseteq \hat{Y}_j$ implies that $c_{ij}$ is a maximum in row $i$.  We can use this observation to establish an even sharper upper bound of $nl_j$: Suppose that $\hat{Y}_j = \bigcup\set{X_{s_1}, \ldots, X_{s_p}}$, and consider the partial granule matrix
\begin{center}
{\small
\begin{tabular}{ccccccc}
& \multicolumn{5}{c}{Decision classes} & \\
Granule in $\hat{Y}_j$  &  $Y_1$ & ... & $Y_j$ & ... & $Y_k$ & Granule size \\ \hline
$X_{s_1}$&  $c_{{s_1}1}$ & ... &  $c_{{s_1}j}$   & ... & $c_{{s_1}k}$    & $c_{s_1}$   \\
... &  ...        & ... &  ...        & ... & ... & ... \\
$X_{s_i}$&  $c_{{s_i}1}$  & ... &  $c_{{s_i}j}$   & ... & $c_{{s_i}k}$    & $c_{s_i}$   \\
... &  ...        & ... &  ...        & ... & ... & ... \\
$X_{s_p}$&  $c_{{s_p}1}$  & ... &  $c_{{s_p}j}$   & ... & $c_{{s_p}k}$    & $c_{s_p}$   \\ \hline
Confusion size  &  $n_{j1}$  & ... &  $n_{jj}$   & ... & $n_{jk}$ &  \\ \hline
\end{tabular}
}
\end{center}
Since a maximum of each row is in column $Y_j$, it follows that $n_{jt} \leq n_{jj}$ for all $1 \leq t \leq k$, and therefore, $\max\set{n_{jt}: 1 \leq t \leq k, t \neq j} \leq n_{jj}$. Setting $nl_j^m \df n_{jj} - \max\set{n_{jt}: 1 \leq t \leq k, t \neq j}\geq nl_j$ we obtain
\begin{theorem}\label{thm:nlmax}
$nl_j \leq nl_j^m \leq nl_j^{**}$ for all $1 \leq j \leq k$.
\end{theorem}

Finally, we estimate the rough upper bound of $Y_j$ using $f_{mrc}$. Setting $nu_j^m \df n_{jj} + \sum_{j\neq i} (n_{ji} + 2 \cdot n_{ij})$, it can be shown that

\begin{theorem}\label{thm:numax}
$nl_j^{**} \leq nu_j^m \leq nu_j$ for all $1 \leq j \leq k$.
\end{theorem}

%\tred{\hrule}

\section{Conclusion and outlook}

In this note, we have explored a connection between rough set approximation and confusion matrices, and have presented several natural indices that approximate the lower and upper bounds given by the reference standard. Owing to lack of space, we have only indicated the procedures with respect to one observer.

The next step will be to broaden the investigation to two or more observers: Each of these has  internal sets $\X$ and $\X'$ of granules which need to be reconciliated to a common standard. This is related to inter--rater reliability which is a common technique used in psychology (and AI) to gauge agreement among experts. We shall also re--interpret common statistics of rough set analysis based on rough confusion matrices. This will, in some sense, complement our earlier research on precision indices in the rough set framework  \cite{gd_standard}.

\section*{Bibliography}

\bibliographystyle{iopart-num}
\bibliography{PRIS2019-P1002}

\end{document}